\documentclass[runningheads]{llncs}
\usepackage[T1]{fontenc}
\usepackage{graphicx}
\usepackage{makecell} 
\usepackage{multirow} 
\usepackage{hyperref} 
\usepackage{amsmath} 
\usepackage{subcaption}  
\usepackage{wrapfig} 

\newcommand{\InitializingVariables}{\textit{Initializing Variables}}
\newcommand{\UpdatingEachSimStep}{\textit{Updating Variables, Each Simulation Step}}
\newcommand{\ConditionalStatements}{\textit{Conditional Statements}}
\newcommand{\UpdatingUnderConditions}{\textit{Updating Variables, Under Conditions}}

\newcommand{\RQone}{How does LC-RAG compare to a discourse-only baseline in terms of the relevance, accuracy, and helpfulness of retrieved knowledge, and how does its performance vary across task context categories?}
\newcommand{\RQtwo}{How do students perceive the epistemic value of Copa's guidance, and to what extent does Copa support their critical thinking?}

\begin{document}
\title{Personalizing Student-Agent Interactions Using Log-Contextualized Retrieval-Augmented Generation (RAG)}

%
\titlerunning{Personalizing Student-Agent Interactions Using Log-Contextualized RAG}
%
\author{
    Clayton Cohn\inst{1}\orcidID{0000-0003-0856-9587} \and
    Surya Rayala\inst{1}\orcidID{0009-0005-8192-8138} \and
    Caitlin Snyder\inst{2}\orcidID{0000-0002-3341-0490} \and
    Joyce Horn Fonteles\inst{1}\orcidID{0000-0001-9862-8960} \and
    Shruti Jain\inst{1}\orcidID{0009-0000-7853-0560} \and
    Naveeduddin Mohammed\inst{1}\orcidID{0000-0002-3706-2884} \and
    Umesh Timalsina\inst{1}\orcidID{0000-0002-5430-3993} \and
    Sarah K. Burriss\inst{1}\orcidID{0000-0002-5598-0363} \and
    Ashwin T S\inst{1}\orcidID{0000-0002-1690-1626} \and
    Namrata Srivastava\inst{1}\orcidID{0000-0003-4194-318X} \and
    Menton Deweese\inst{3}\orcidID{0000-0001-7361-3826} \and
    Angela Eeds\inst{3} \and
    Gautam Biswas\inst{1}\orcidID{0000-0002-2752-3878}
}
%
\authorrunning{C. Cohn et al.}
%
\institute{
    \textsuperscript{1}Department of Computer Science, Vanderbilt University, Nashville, USA \\
    \textsuperscript{2}College of Engineering \& Science, University of Detroit Mercy, Detroit, USA \\
    \textsuperscript{3}The School for Science and Math, Vanderbilt University, Nashville, USA \\
    \email{clayton.a.cohn@vanderbilt.edu}
}
\maketitle              
\begin{abstract}
Collaborative dialogue offers rich insights into students' learning and critical thinking, which is essential for personalizing pedagogical agent interactions in STEM+C settings. While large language models (LLMs) facilitate dynamic pedagogical interactions, \textit{hallucinations} undermine confidence, trust, and instructional value. Retrieval-augmented generation (RAG) grounds LLM outputs in curated knowledge, but requires a clear semantic link between user input and a knowledge base, which is often weak in student dialogue. We propose log-contextualized RAG (LC-RAG), which enhances RAG retrieval by using environment logs to contextualize collaborative discourse. Our findings show that LC-RAG improves retrieval over a discourse-only baseline and enables our collaborative peer agent, Copa, to deliver relevant, personalized guidance that supports students' critical thinking and epistemic decision-making in the collaborative computational modeling environment C2STEM.

\keywords{NLP \and LLMs \and RAG \and Agents}

\end{abstract}

\section{Introduction} \label{sec:intro}

The integration of generative AI (GenAI) into collaborative learning environments raises important questions about how this technology can best help students construct knowledge, make decisions, and engage in shared problem-solving activities. Collaborative computational modeling --- where students use code to construct models that simulate and analyze scientific processes \cite{snyder2019analyzing} --- enhances science, technology, engineering, mathematics, and computing (STEM+C) learning \cite{hutchins2020c2stem}. However, these complex environments can also create problem-solving challenges \cite{cohn2024towards}, highlighting the need for adaptive supports that align with students' evolving understanding and promote epistemic agency.

Vygotsky's \textit{zone of proximal development} (ZPD) \cite{vygotsky1978mind} emphasizes guided support as paramount in helping learners accomplish tasks they cannot yet complete independently \cite{cohn2025exploring}. Pedagogical agents help bridge this gap by guiding individual learners and facilitating collaboration \cite{cohn2025exploring,stamper2024enhancing}. However, developing agents to support collaborative STEM+C problem-solving is non-trivial, as agents must adapt to both individual learners and groups while guiding learning and collaboration across multiple domains simultaneously (e.g., physics and computing). 

Collaborative dialogue provides insights into students' learning and critical thinking, helping identify knowledge gaps and problem-solving approaches \cite{snyder2019analyzing}. This information can improve pedagogical agents, which struggle to (1) ground their interactions in ongoing problem-solving discussions; and (2) offer dynamic and personalized adaptive scaffolding beyond rigid, rule-based approaches. However, dialogue alone is insufficient for effectively guiding students through challenges, as many students struggle with help-seeking and often fail to recognize when they need assistance \cite{roll2014benefits}. 

Jurenka et al. (2024) \cite{jurenka2024towards} emphasize that pedagogical agents should ``see what the student sees,'' highlighting the need for agents that are aware of students' actions inside the learning environment. Previous research has shown that integrating collaborative discourse with learning environment activity logs can provide a more comprehensive understanding of student learning \cite{cohn2024towards,snyder2024analyzing,snyder2025using}. In this context, large language models (LLMs) offer a promising solution \cite{stamper2024enhancing}.

While LLMs have been applied across various artificial intelligence in education (AIED) tasks like intelligent tutoring \cite{stamper2024enhancing} and formative assessment scoring \cite{cochran2023_AIED_improving,cochran2022improving,cohn2024chain}, they are prone to \textit{hallucinations} that lead to errors and undermine trust in educational systems \cite{cohn2025cotal}. Fine-tuning can help reduce hallucinations, but it typically requires substantial amounts of data and computational resources and may limit LLMs' ability to generalize beyond their training data \cite{gekhman2024does}.

Recently, retrieval-augmented generation (RAG \cite{lewis2020retrieval}) has emerged as an effective alternative to fine-tuning, allowing LLM agents to dynamically access information from a knowledge base during inference without additional training \cite{yan2024vizchat}. Domain knowledge --- curated by humans --- is chunked, transformed into vector embeddings that capture semantic meaning \cite{yan2024vizchat}, and stored in a vector database (i.e., knowledge base). When a user inputs a query, it is embedded and matched to relevant knowledge base embeddings via semantic search, then retrieved to facilitate personalized, context-aware LLM responses. 

However, effective RAG retrieval relies on semantic alignment between user input and the knowledge base, often measured by cosine similarity \cite{yan2024vizchat,lewis2020retrieval}. Without this, agents risk retrieving irrelevant information, leading to unhelpful responses. While RAG is well-suited for question-answering tasks \cite{qian2024memorag}, where queries align closely with the knowledge base, it faces challenges in collaborative discussions where students may struggle to express the information they need. To improve RAG-based interventions in collaborative settings, mechanisms are needed to link student conversations to pertinent knowledge base information.

In this paper, we address the semantic gap between collaborative dialogue and knowledge base content by leveraging environment log data to enhance RAG retrieval, improving agent interactions as high school students problem-solve in the collaborative computational modeling environment, C2STEM. Our method, log-contextualized retrieval-augmented generation (LC-RAG), integrates collaborative discourse with environment logs by using an LLM to generate summaries of student learning across problem-solving segments, and aligns with a knowledge base of physics and computing textbooks to improve RAG retrieval. 

We evaluate LC-RAG's performance across multiple embedding models and sizes, as well as four \textit{task context} categories \cite{snyder2025using} (described in Section \ref{subsec:experimental_design}) derived from C2STEM environment logs based on the specific computational model components students engage with. Additionally, we introduce an agent powered by LC-RAG and GPT-4o, Copa (\textbf{Co}llaborative \textbf{P}eer \textbf{A}gent), and conduct a classroom study and focus groups with students to assess the agent's epistemic value, aiming to answer the following Research Questions:

\begin{enumerate}
    \item \textbf{RQ1}: \RQone
    \item \textbf{RQ2}: \RQtwo
\end{enumerate}

Our findings show that (1) LC-RAG improves RAG retrieval, enabling Copa to provide personalized support that fosters critical thinking; and (2) students perceive their interactions with Copa as epistemically valuable. This research advances AIED methodology by improving the ability of pedagogical agents to interpret collaborative discourse and deliver context-aware interventions in computational modeling environments.

\section{Background} \label{sec:background}

RAG is valuable for aligning LLM outputs with educator preferences, giving teachers greater control over responses. However, it requires a semantic link between input text and the knowledge base, which is often missing in collaborative discourse. While several researchers have identified this disconnect, few have explored solutions. 

The \textit{MemoRAG} framework \cite{qian2024memorag} utilizes a lightweight LLM as a memory mechanism to create a compressed representation of the knowledge base, generating `clues' from user input to improve knowledge base alignment. However, training such a memory module can be impractical in educational settings due to high computational demands and limited data \cite{cochran2023_CSEDU_improving}.

Yan et al. (2024) \cite{yan2024vizchat} incorporate RAG into a learning analytics dashboard for nursing students to facilitate reflection during post-simulation debriefings. Their system encourages students to clarify ambiguous queries to improve semantic matches within the knowledge base. While this approach is suitable for adults in post hoc settings, it may not be effective for children who struggle to articulate their needs in real time. Furthermore, interrupting collaborative discussions for clarification can disrupt learning and diminish user experience. 

LlamaIndex's Query Transformations module \cite{llamaindex_query_transformations} enables users to refine queries by generating a hypothetical document or answer for RAG retrieval. However, this assumes that the initial query is semantically relevant to the task and the knowledge base, which is often not the case in collaborative discourse due to knowledge gaps and off-topic conversations. A notable research gap remains in developing RAG methods that effectively retrieve relevant knowledge during collaborative dialogue, which we address by leveraging log-contextualized summarization with LC-RAG.

\section{Methods} \label{sec:methods}

\subsection{LC-RAG} \label{subsec:lc_rag}
\begin{figure}[htbb]
    \centering
    \includegraphics[width=1\linewidth]{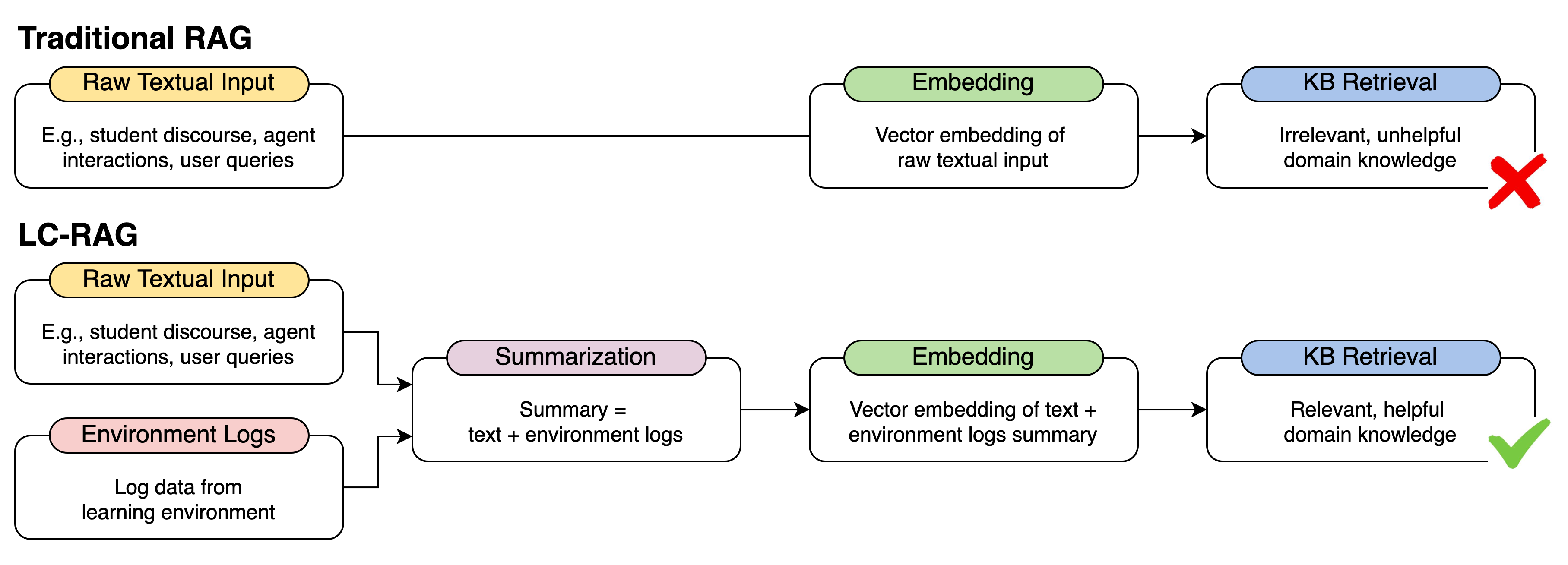}
    \caption{Traditional RAG (top) vs. LC-RAG (bottom).}
    \label{fig:lc_rag}
\end{figure}

LC-RAG augments RAG retrieval by integrating student interactions and environment log data for use in pedagogical agents. Instead of using raw text for semantic search, LC-RAG uses an LLM to generate summaries that integrate discourse and log data, thereby contextualizing student interactions. These summary embeddings are mapped to the knowledge base, enabling contextualized domain knowledge retrieval during LLM inference. Figure \ref{fig:lc_rag} illustrates this process. LC-RAG supports students’ epistemic agency by grounding agent responses in contextually relevant information, helping them construct, validate, and apply disciplinary knowledge during collaborative problem-solving. Section \ref{subsec:experimental_design} discusses the design and implementation of LC-RAG in this paper.

\subsection{C2STEM Learning Environment} \label{subsec:c2stem}

\begin{figure}[htbp]
    \centering
    \includegraphics[width=1\linewidth]{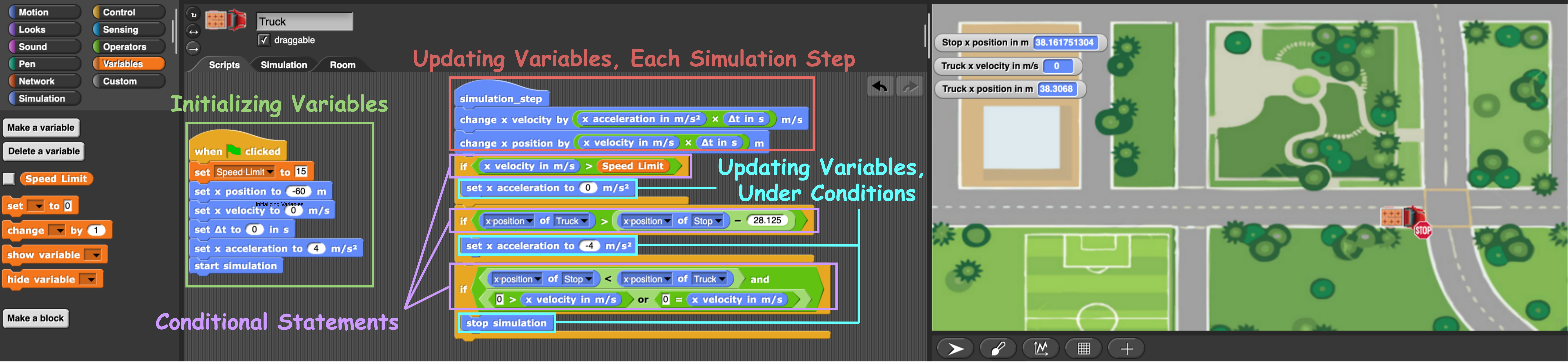}
    \caption{C2STEM Truck Task example solution with task context categories.}
    \label{fig:c2stem}
\end{figure}

C2STEM \cite{hutchins2020c2stem} is a collaborative computational modeling environment targeting physics and computing where students create and debug computational models simulating scientific phenomena (shown in Figure \ref{fig:c2stem}). In this study, students engaged in a one-dimensional \textit{Truck Task} activity in which they modeled a truck's motion, accelerating from rest, cruising, and decelerating to stop at a stop sign.  

\subsection{Experimental Design} \label{subsec:experimental_design}
All studies described below were conducted in Nashville, TN, USA with approval from Vanderbilt University's Institutional Review Board (IRB), and all participating students and their parents provided assent and consent, respectively. 

\subsubsection{RQ1 (LC-RAG)} \label{subsubsec:rq1_design}
To answer RQ1, we conducted an eight-week study with 24 high school sophomores, ages 15-16, participating in a kinematics curriculum with one two-hour session each week. Students worked in pairs, sharing a laptop. We collected (and transcribed) audio and environmental log data from 18 consented students during the Truck Task, yielding 2,786 spoken utterances and 2,275 logged actions over 9 hours of collaborative discourse.  

We used environment logs to segment the multimodal data based on the model components students worked on. Using an abstract syntax tree (AST), we classified logged student actions into four \textit{task context} categories \cite{snyder2025using}: (1) \InitializingVariables, (2) \UpdatingEachSimStep, (3) \ConditionalStatements, and (4) \UpdatingUnderConditions\ (see Figure \ref{fig:c2stem}). For instance, blocks placed under ``When Green Flag Clicked'' were labeled as \InitializingVariables\ segments. This method ensured that student discourse reflected their problem-solving context. We identified a total of \textbf{n=216} problem-solving segments, with \ConditionalStatements\ being the most common (77 segments, average discourse length of 790 characters) and \UpdatingEachSimStep\ the least common (36 segments, 443 characters).

We employed GPT-3.5-Turbo, selected for its balance of performance and cost, to summarize each segment based on its discourse and task context. For instance, when a student said, \textit{``put that position block there,''} task context helped the LLM infer that the students were setting the truck's \textit{initial} position because that segment's task context was \InitializingVariables. These summaries were used by LC-RAG to retrieve domain knowledge, replacing raw discourse that lacked context. Our summarization prompt, provided in the \href{https://github.com/claytoncohn/AIED25_Supplementary_Materials}{Supplementary Materials}\footnote{https://github.com/claytoncohn/AIED25\_Supplementary\_Materials}, was extensively tested. While certain summary words and characters varied across runs, semantic meaning remained intact. Two researchers validated each summary against the raw data, ensuring accuracy before analysis.

We evaluated LC-RAG's ability to retrieve relevant information from a knowledge base relative to a \textit{discourse-only baseline} (i.e., without using environment logs). Our analysis involved five embedding models of different dimensions (discussed shortly) to assess LC-RAG's performance in terms of relevance, accuracy, and helpfulness across the 216 student problem-solving segments. For our knowledge base, we selected two high school textbooks --- one for each domain in C2STEM, physics \cite{urone2020physics} and computing \cite{bourke2018computer} --- due to their open-source availability, curricular relevance, and age appropriateness. 

For each of the 216 segments, we generated two embeddings: one using only the student discourse (baseline) and one using the LLM-generated segment summary that incorporates discourse and task context (LC-RAG). To assess robustness, we tested five embedding models with varying dimensions: openai-text-embedding-3-large-3072, openai-text-embedding-3-small-1536, voyage-3-large-1024, voyage-3-lite-512, and microsoft-e5-large-1024. OpenAI (OAI) models were chosen for their high performance and ubiquity, Voyage models for their enterprise adoption, and the Microsoft model as a lightweight, open-source alternative. 

For each model-dimension combination, we used semantic search with cosine similarity \cite{yan2024vizchat} to retrieve the most relevant domain knowledge chunk from the knowledge base. This produced 2,160 retrieved chunks: 5 model-dimension combinations $\times$ 216 segments $\times$ 2 segment embeddings. Due to the lack of ground-truth labels and the scale of the evaluation, we employed an \textit{LLM-as-a-Judge} \cite{zheng2023judging} protocol using GPT-4o to compare LC-RAG against the baseline in terms of relevance, accuracy, and helpfulness. This approach, which reports \textit{win rates} rather than traditional metrics such as F1, precision, or recall, is commonly used in similar evaluation settings \cite{zheng2023judging}. GPT-4o was selected for its state-of-the-art performance.

For each of the 1,080 comparisons (5 model-dimension combinations $\times$ 216 segments), GPT-4o anonymously compared LC-RAG's retrieved chunk to the baseline's and determined which would lead to a more relevant, accurate, and helpful LLM generation, considering the C2STEM Truck Task (see Section \ref{subsec:c2stem}) and student discourse for that segment. The model was instructed to justify its decision using step-by-step \textit{chain-of-thought} reasoning \cite{wei2022chain} to aid analysis. Following the original LLM-as-a-Judge protocol, we used \textit{win rate} as our evaluation metric, where LC-RAG's win rate for each embedding model-dimension combination was calculated as $lc\_rag\_wins / 216$. We additionally report loss and tie rates for a more comprehensive comparison.

While previous AIED research has effectively used LLM-as-a-Judge for evaluation \cite{parker2024large}, studies show that variations in prompt phrasing \cite{mizrahi2024state} and example ordering \cite{lu2021fantastically} can significantly impact outcomes. To enhance robustness, Mizrahi et al. (2024) \cite{mizrahi2024state} recommend multi-prompt evaluation, aggregating results across multiple templates. Accordingly, we created three separate evaluation prompts and aggregated the judges' predictions by majority vote. Chunk order was randomized in each run, following Lu et al. (2021) \cite{lu2021fantastically}. All three evaluation prompts are available in our \href{https://github.com/claytoncohn/AIED25_Supplementary_Materials}{Supplementary Materials}. 

To validate our LLM-as-a-Judge protocol, we randomly sampled 80 segments across all model-dimension combinations. Two authors scored all 80 retrieval pairs via consensus coding and compared LC-RAG's retrievals to the baseline's using the same criteria given to the LLM. We assessed inter-rater reliability (IRR) using Cohen's Quadratic Weighted Kappa (QWK) because the data were ordinal, yielding QWK$=0.49$ (moderate agreement).

Each of the four task context categories discussed previously requires different domain knowledge. For example, \InitializingVariables\ primarily involves initializing constants and variables (computing), while \UpdatingEachSimStep\ requires an understanding of loops (computing) and kinematic variable relationships (physics). This variation necessitated evaluating LC-RAG's performance across these contexts, which we addressed by averaging the win rates of LC-RAG and the baseline across the five embedding models for all four task context categories to identify trends. For RQ1, we sampled LLM responses across all three prompt templates for qualitative analysis, thereby enriching our quantitative findings with deeper insights.

\subsubsection{RQ2 (Copa)} \label{subsubsec:rq2_design}

To answer RQ2, we conducted two studies with a total of \textbf{n=18} students interacting with Copa: (1) a one-day, 2.5-hour classroom study with 16 high school freshmen (ages 14–15), where students solved a simplified version of the Truck Task (excluding the deceleration and stopping conditions); and (2) two one-hour focus groups with six students --- four from the classroom study and two from the earlier study for RQ1 --- who debugged a fictional student's faulty Truck Task model using Copa.

Classroom study participants took part in the research as part of their regular curriculum. The study began with an instructional session introducing students to C2STEM, followed by the Truck Task problem-solving activity using Copa. After the activity, one of the authors of this paper facilitated an interactive class discussion using a Miro storyboarding exercise to reflect on Copa's accuracy, helpfulness, and trustworthiness. The session concluded with a semi-structured discussion, led by another author, exploring students' broader perceptions of Copa. Data collected included student conversations, environment actions, agent interactions, screen recordings, Miro board notes, and discussion notes.

Focus group participants volunteered to participate outside of regular class time and received a \$25 Amazon gift card for their involvement. Students were first given task instructions explaining that they would use Copa to debug a fictional student's erroneous C2STEM Truck Task code. After completing the debugging activity, one of this paper's authors led a semi-structured discussion to gather students' perceptions of Copa. Participants also completed a survey about their experience, including their likes and dislikes as well as their views on the accuracy, helpfulness, and trustworthiness of the agent's feedback. Data collected included student conversations, webcam recordings, environment actions, agent interactions, screen recordings, discussion notes, and survey responses.

To determine Copa's epistemic value to students and its ability to promote critical thinking, we conducted a qualitative analysis by reviewing students' interactions with Copa (including retrieved domain knowledge), Miro boards, discussion notes, and focus group survey responses, memoing key findings across all sources \cite{hatch2002} and aggregating the results. Results for both RQs are presented in the next section.

\section{Results} \label{sec:results}

\subsection{LC-RAG (RQ1)} \label{subsec:findings_lc_rag}

We answer RQ1 by comparing LC-RAG's retrieval performance to the baseline's across five embedding model-dimension combinations (see Section \ref{subsubsec:rq1_design}). Results are presented in Table \ref{tab:results_metrics}.

\begin{table}[htbp]
    \centering
    \begin{tabular}{|c|c|c|c|c|c|}
        \hline
        \textbf{Model} & \makecell{\textbf{Emb.} \\ \textbf{Size}} & \makecell{\textbf{Loss} \\ \textbf{Rate}} & \makecell{\textbf{Tie} \\ \textbf{Rate}} & \makecell{\textbf{Win} \\ \textbf{Rate}} \\
        \hline
        \multicolumn{1}{|l|}{OAI-text-embedding-3-large} & 3072 & 0.343 & 0.255 & \textbf{0.403} \\
        \hline
        \multicolumn{1}{|l|}{OAI-text-embedding-3-small} & 1536 & \textbf{0.690} & 0.093 & 0.218 \\
        \hline
        \multicolumn{1}{|l|}{voyage-3-large} & 1024 & 0.329 & 0.227 & \textbf{0.444} \\
        \hline
        \multicolumn{1}{|l|}{voyage-3-lite} & 512 & 0.347 & 0.171 & \textbf{0.482} \\ 
        \hline
        \multicolumn{1}{|l|}{microsoft-e5-large} & 1024 & 0.213 & 0.167 & \textbf{0.620} \\ \hline
    \end{tabular}
    \vspace*{2mm} 
    \caption{LC-RAG \textbf{Win}, \textbf{Loss}, and \textbf{Tie} rates relative to the baseline. \textbf{Model} refers to embedding model, and \textbf{Emb. Size} refers to embedding dimensionality.}
    \label{tab:results_metrics}
\end{table}

\textbf{LC-RAG outperformed the baseline in 4 out of 5 embedding model-dimension combinations}, achieving more precise retrieval by isolating a smaller set of relevant chunks compared to the baseline. LC-RAG retrieved 167 unique chunks from a knowledge base of 7,386, while the baseline retrieved 267. This was advantageous, particularly when student discussions were off-topic and resulted in irrelevant retrieval for the baseline. For example, consider a segment where students said, ``\textit{S1: We need help. S2: No we don't. We got this. We can do this.}'' In this case, using the voyage-3-lite embedding model, the baseline retrieved an irrelevant chunk about a `90s TV show (``\textit{...from an episode of Ren and Stimpy...}''), while LC-RAG retrieved a chunk related to scientific modeling.

The baseline outperformed LC-RAG with OpenAI's ``small'' embedding model because the knowledge base was general in nature, comprising physics and computing domain knowledge outside the scope of the C2STEM Truck Task (e.g., thermodynamics and memory management). The baseline's raw discourse often lacked references to the C2STEM environment or Truck Task, whereas LC-RAG's integrated discourse and log summaries (see Section \ref{subsec:lc_rag}) reframed student conversations within the context of C2STEM, the Truck Task, and their environment actions. While this contextualization made LC-RAG summaries more meaningful within C2STEM, it hindered retrieval performance, as the knowledge base lacked C2STEM or Truck Task information. Although the baseline outperformed LC-RAG quantitatively, qualitative analysis revealed that both models were prone to irrelevant retrievals, suggesting that the limitation stems from the untargeted knowledge base rather than the retrieval mechanism itself. We address this while answering RQ2.

\begin{wrapfigure}{r}{0.5\textwidth}
  \begin{center}
    \includegraphics[width=0.48\textwidth]{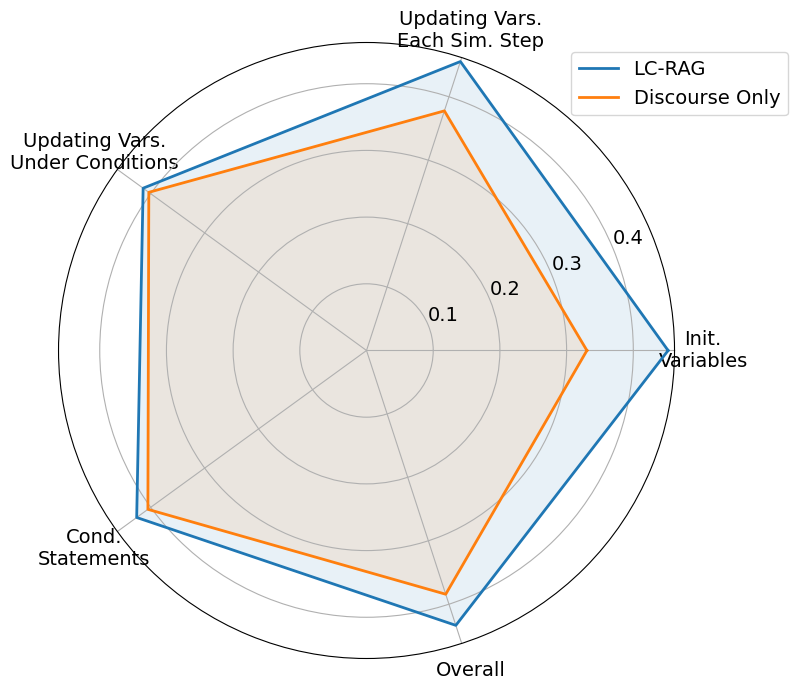}
  \end{center}
  \caption{LC-RAG win rates by task context category averaged over all five embedding models. \textit{Overall} compares performance across all task contexts.}
  \label{fig:task_context}
\end{wrapfigure}

Across all four task context categories (and overall), LC-RAG achieved a higher win rate than the baseline when considering results from all five embedding models. LC-RAG's outperformance was more pronounced in \InitializingVariables\ (45\% for LC-RAG, 33\% for the baseline) and \UpdatingEachSimStep\ (46\% for LC-RAG, 38\% for the baseline) segments but less pronounced for \ConditionalStatements\ and \UpdatingUnderConditions, differing only by 1-2\%. 

Performance gains were reduced for \UpdatingUnderConditions\ and \ConditionalStatements\ for the same reason as previously discussed: when using the OpenAI ``small'' model, the LC-RAG summaries were too task-specific to align well with the generalized knowledge base. Excluding this model, all four task-context categories showed clear improvement over the baseline. This underperformance, while limited to a single embedding model, highlights a limitation in how our task context categories contextualize student discourse. Our RQ2 analysis explores an alternative approach to integrating log data that generates segment summaries with stronger semantic links to the knowledge base.

\subsection{Copa (RQ2)} \label{subsec:findings_copa}

We answer RQ2 by qualitatively analyzing agent interactions, Miro boards, class discussions, and survey responses from the classroom and focus group studies (see Section \ref{subsubsec:rq2_design}). Building on findings from RQ1, we refined LC-RAG in two ways. First, we used a condensed knowledge base comprising 15 task-specific chunks (e.g., kinematic equations, computational concepts such as initializing and updating variables). Second, we adopted an \textit{agentic} approach (i.e., allowing the LLM to analyze data, reason, and act) to integrate log and text data. Rather than relying on predefined task context categories, we provided the LLM with the current state of the students’ computational model. During summarization, the LLM identified the problem students expressed, diagnosed its cause by comparing the student model to an expert version, and recommended relevant domain knowledge to retrieve. This recommendation was then embedded for semantic search. Copa’s summarization and interface prompts were iteratively refined by this paper's authors and are provided in the \href{https://github.com/claytoncohn/AIED25_Supplementary_Materials}{Supplementary Materials}.

Overall, students found Copa to be epistemically valuable. All focus group participants reported that their interactions with Copa supported their problem-solving and the agent's information was accurate. This sentiment was echoed in classroom discussions and on the Miro boards, where many students emphasized how Copa helped them better understand the course material (``\textit{It felt like Copa was in the room with us the entire time trying to teach us}'';``\textit{Has a step-by-step process that is easy to follow.}'';``\textit{Provided a reflection after we asked the questions}''). Several students also highlighted Copa’s positive tone as a source of motivation and encouragement (``\textit{...polite and encouraging feedback}'',``\textit{... praising us when we got a correct answer}'',``\textit{My favorite part about COPA was its motivative language}''). One student, who had struggled with the C2STEM Truck Task two years earlier when no agent was available, reported feeling ``\textit{a lot more confident}'' and ``\textit{got a lot further on [her] own this time around}.'' 

Copa supported students' critical thinking and epistemic decision-making (``\textit{...helped me critically think...}''), enabling them to grasp domain concepts they had previously struggled to understand. In one instance, a group admitted to Copa that they ``\textit{...have no idea...}'' how to calculate the lookahead distance, i.e., the distance required for the truck to decelerate and stop at the stop sign. Recognizing their need for guidance, Copa retrieved and presented a relevant kinematic equation from the knowledge base using LC-RAG. With this information, the students realized they needed to calculate the truck’s displacement, gaining a clearer understanding of its role in both the equation and the deceleration phase of the Truck Task. 

However, several students expressed frustration when Copa did not provide direct answers (``\textit{it took a while to give us what we wanted}'',``\textit{I would like it to give us a straight answer}'') and were unable to think critically about Copa's suggestions. Copa is designed as a peer agent that collaboratively problem-solves alongside students, offering suggestions that may be incorrect and are intended to prompt analysis and evaluation. In some cases, students assumed Copa’s recommendations were correct and became frustrated when those suggestions failed to improve their models (``\textit{Copa suggested us to try something that ended up making the code worse.}''; ``\textit{...it did not give correct information}''). Although the lead author of this paper encouraged students to think critically about Copa’s responses—acknowledging that Copa might not always be right—some still became upset when its guidance proved unhelpful and resorted to trial-and-error approaches rather than deepening their understanding of the problem. This highlights a fundamental trade-off between students’ epistemic agency and teachers' curricular goals, which we explore further in Section \ref{sec:discussion_conclusion}.

Despite these frustrations, students consistently emphasized that Copa was epistemically valuable. All focus group participants reported that they would trust Copa to support them in future problem-solving activities. Several students specifically cited its conceptual understanding and awareness of their actions in the C2STEM environment as reasons for this trust (``\textit{...it knew what we were doing/did}''). LC-RAG effectively retrieved domain knowledge that was both relevant and helpful, enabling personalized agent interactions that advanced students’ task goals. For instance, when one group asked, ``\textit{how should I expand the ``Simulation\_step'' block}'' --- a query with no direct semantic match in the knowledge base --- LC-RAG successfully retrieved kinematic equations used to update velocity and position, which are the first two steps under that block.

\section{Discussion and Conclusions} \label{sec:discussion_conclusion}

Our findings highlight important implications for AIED theory, methodology, and practice. Theoretically, LC-RAG aligns with Vygotsky’s ZPD \cite{vygotsky1978mind} by enabling context-aware scaffolding that adapts to students’ evolving needs. By integrating discourse with environment logs, LC-RAG allows pedagogical agents to identify where students are within their ZPD and deliver timely, personalized support. This dynamic scaffolding moves beyond static, rule-based systems, enabling agents to respond meaningfully to students’ immediate challenges.

Methodologically, our RQ2 findings suggest that agentic reasoning enhances personalized student support. By \textit{characterizing} students' difficulties and \textit{recommending} domain knowledge, based on students’ verbalized needs and their actions within the C2STEM environment, Copa generates embeddings that align closely with the knowledge base, enabling effective RAG retrieval and more contextualized interactions. This process introduces a transparent reasoning chain: model outputs are grounded in retrieved content, which is tied to the LLM’s reasoning and can be traced back to students' actions and the problems they express. This structure not only supports effective personalization but also enhances explainability, offering insight into both what the agent recommends and why. We hypothesize that (1) additional conversational agent components (e.g., dialogue managers), could similarly benefit from agentic reasoning; and (2) agentic reasoning will allow LLM-based agents to support students' learning processes by deepening their conceptual understanding and promoting epistemic growth over time, positioning agentic AIED as a compelling direction for future research.

As discussed in Section \ref{subsec:findings_copa}, students’ desire for direct answers --- rather than engaging with domain concepts through problem-solving --- reveals a core tension between students’ epistemic agency and teachers’ goals for deeper conceptual understanding. Even when students struggle, prior work has emphasized the value of productive failure in promoting long-term understanding \cite{cohn2024multimodal}. As students become more accustomed to tools such as ChatGPT, their expectations may increasingly favor immediate answers, fostering over-reliance on GenAI and frustration when they are encouraged to think critically while working toward a solution. Research on the role design of pedagogical agents has found that while teachers and students often agree on feedback principles, students tend to prefer feedback that directly addresses task completion, whereas teachers favor feedback that supports broader learning and collaboration goals \cite{cohn2025exploring}.

This raises an important question: how much epistemic agency should pedagogical agents grant students if that agency risks fostering over-reliance and hindering learning? We argue that while students should exercise some degree of epistemic agency in their interactions with agents, this agency should not come at the expense of opportunities for critical thinking. Beyond supporting learning and collaboration, GenAI-powered agents should treat problem-solving as a valuable skill in its own right --- one worth cultivating independently of any specific task. Research on self-regulated learning has shown a strong link between students’ self-regulation and critical thinking abilities \cite{kusmaryono2023critical}. To support this, GenAI agents must be grounded in learning theories whose core pedagogical principles remain intact, even as they facilitate agency.

More broadly, this underscores the importance of grounding LLM-powered agents in established learning theories. The accessibility and ease of use of LLMs have led to a surge in pedagogical agents often deployed without a clear theoretical foundation --- marking a departure from pre-LLM intelligent tutoring systems, which were tightly aligned with cognitive and learning science principles. Recent AIED research has raised concerns about the risks of designing LLM systems that overlook the pedagogical foundations of agent feedback \cite{stamper2024enhancing}. As LLMs become increasingly central to AIED, theory-driven design is essential to ensure these systems do more than mimic helpfulness --- they must reflect how students actually learn. Grounding intelligent systems in learning theory is also critical for building trust among practitioners, who require confidence that AI interventions --- particularly GenAI ones --- are pedagogically sound, not just technologically advanced.

Our work is not without limitations. Further research is needed to assess (1) how LC-RAG generalizes across different learning environments, agent architectures, and subject domains; and (2) Copa's impact on students' behaviors and learning outcomes. LC-RAG also relies on the ability to represent environment log data in a canonical, semantically meaningful form that is accessible to LLMs, which may not always be feasible. Additionally, Copa was not evaluated in a controlled study, so we cannot directly compare LC-RAG’s contribution to its performance with that of other RAG approaches. 

Future research will explore multi-agent architectures where specialized agents work together to enhance retrieval, make pedagogical decisions, and generate feedback, ultimately improving agent reasoning and the quality of support provided. We will also improve LC-RAG's grounding by integrating additional modalities, such as student-agent interactions, self-regulation indicators, and affective signals. This approach aims to provide a more comprehensive understanding of students' evolving problem-solving processes. We will evaluate this strategy through controlled and longitudinal studies to assess its impact on retrieval quality, interaction patterns, critical thinking, and learning over time.

Overall, this paper demonstrates that integrating student verbalizations and environment logs enables pedagogical agents to support rich, multi-turn interactions that foster critical thinking and are perceived by students as epistemically valuable. We aim to establish a foundation for future research on adaptive, context-aware GenAI agents for personalized learning in collaborative STEM+C environments and beyond.


%
%
%
\bibliographystyle{splncs04}
\bibliography{references}

\end{document}